\documentclass{article}

\usepackage{amsmath, amsfonts, amsthm, amssymb, mathtools, xcolor, bbm, parskip}
\usepackage{mymacros}

\usepackage{fontawesome5}

\usepackage[multiple]{footmisc}

\usepackage{tikz, ifthen}

\mathtoolsset{showonlyrefs,showmanualtags}

\usepackage[hidelinks, hyperfootnotes=false]{hyperref}
\usepackage{xurl}

\newcommand{\DKL}{D_{\mathrm{KL}}}

\usetikzlibrary{positioning, calc}

\tikzstyle{neuron} = [circle,draw, minimum size=.9cm, inner sep=0]

\title{A Concise Mathematical Description of Active Inference in Discrete Time}
\author{Jesse van Oostrum, Carlotta Langer, Nihat Ay}
\date{}

\usepackage{biblatex} 
\usepackage{xurl}

\begin{document}

\maketitle

\begin{abstract}
    In this paper we present a concise mathematical description of active inference in discrete time. The main part of the paper serves as a basic introduction to the topic, including a detailed example of the action selection mechanism. The appendix discusses the more subtle mathematical details, targeting readers who have already studied the active inference literature but struggle to make sense of the mathematical details and derivations. Throughout, we emphasize precise and standard mathematical notation, ensuring consistency with existing texts and linking all equations to widely used references on active inference. Additionally, we provide Python code that implements the action selection and learning mechanisms described in this paper and is compatible with \texttt{pymdp} environments. 
\end{abstract}

\section*{Introduction}
Active inference is a theory that describes the action selection and learning mechanisms of an agent in an environment. We aim to present a concise mathematical description of the theory so that readers interested in the mathematical details can quickly find what they are looking for. We have paid special attention to choosing notation that is more in line with standard mathematical texts and is also descriptive, in the sense that dependencies are made explicit. Hence, the focus of this paper lies on the mathematical details and derivations rather than verbal motivations and justifications.

The paper consists of a main text and an appendix. The main text provides a clear introduction to active inference in discrete time, accessible for people new to the topic. It is divided into two parts: inference, which assumes a given generative model, and learning, which explains how the agent acquires this model. The main text concludes with a worked example of action selection. The appendix delves into finer details and derivations, catering to readers familiar with active inference who seek clarity on the mathematical aspects.

To complement our theoretical exposition, we provide a Python implementation\footnote{\url{https://github.com/jessevoostrum/active-inference}} of the action selection and learning mechanisms described in this paper, which is compatible with \texttt{pymdp} environments. This code is more minimalistic, which makes it easier to understand than other implementations such as \texttt{SPM} and \texttt{pymdp}.

\section{Set-up and notation} \label{sec:notation}

In this paper we consider an active inference agent acting in a discrete-time setting with a finite-time horizon. This means that we consider a sequence of $T$ time steps and at every time step $\tau$ the agent receives an observation $o_\tau$, and performs an action $a_\tau$. We use $\tau$ for arbitrary time steps and the letter $t$ to denote the current time step. We use the subscript ${}_{\tau:\tau'}$ to denote a sequence of variables, e.g.\ $o_{\tau:\tau'} = (o_\tau, \ldots, o_{\tau'})$. A sequence of (future) actions is called a \emph{policy}\footnote{Note that in a reinforcement learning context the term ``policy" has a different meaning.} and is denoted by $\pi_t = a_{t:T}$, with $\pi = \pi_1$. We write $a_{1:t-1}$ for actions that were performed in the past and $\pi_t$ for future actions that still need to be selected. 

\begin{figure} [ht]
    \centering
    \begin{tikzpicture}
        \pgfmathsetmacro{\width}{2}
        \pgfmathsetmacro{\heighto}{-1.5}
        \pgfmathsetmacro{\heighta}{1.3}

        % \node (pi) at (.85, 1.7) [neuron] {$\pi$};
        \node (s1) at (0,0) [neuron] {$s_1$};
        \node (s2) at (1*\width,0) [neuron] {$s_2$};
        \node (s3) at (2*\width,0) [neuron] {$s_3$};
        \node (o1) at (0,\heighto) [neuron] {$o_1$};
        \node (o2) at (1*\width,\heighto) [neuron] {$o_2$};
        \node (o3) at (2*\width,\heighto) [neuron] {$o_3$};
        \node (a1) at (.5*\width,\heighta) [neuron] {$a_1$};
        \node (a2) at (1.5*\width,\heighta) [neuron] {$a_2$};

        \draw[->] (s1) -- (s2); 
        \draw[->] (s2) -- (s3); 
        \draw[->] (s1) -- (o1); 
        \draw[->] (s2) -- (o2); 
        \draw[->] (s3) -- (o3); 
        \draw[->] (a1) -- (s2); 
        \draw[->] (a2) -- (s3); 

        \node (sT) at (3*\width + .5,0) [neuron] {$s_T$};
        \node (oT) at (3*\width + .5,\heighto) [neuron] {$o_T$};
        \node (aT-1) at (2.5*\width + .5,\heighta) [neuron] {$a_{T-1}$};

        \draw[->] (sT) -- (oT); 
        \draw[->] (aT-1) -- (sT); 

        \draw[->, dotted] (s3) -- (sT);

    \end{tikzpicture}
    \caption{Graphical representation of the generative model}
    \label{fig:representation-genmod}
\end{figure}
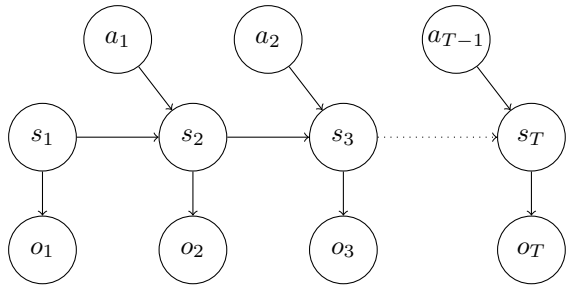

The agent models the dynamics of the environment using an internal generative model. This model uses a variable $s_\tau$, called an internal state, to represent the state of the environment\footnote{Note that the number of possible internal states is usually much smaller than the actual number of states the environment can be in.} at time step $\tau$. The model is given by the following probability distribution:
\begin{align}
    p(o_{1:T}, s_{1:T} | a_{1:T-1}, \theta).
\end{align}
This probability distribution factorizes according to the graph in Figure \ref{fig:representation-genmod}.
In the first part of this paper we assume that this generative model is given and need not be learned, and we therefore suppress the dependence on the parameter $\theta$. In the second part we discuss how the model is learned. 

Suppose the agent is at time step $t$. It will have received observations $o_{1:t}$ and performed actions $a_{1:t-1}$. We use $q_t(s_{\tau:\tau'})$ to denote the (approximate) posterior distribution of the generative model, $p( s_{\tau:\tau'}| o_{1:t}, a_{1:t-1})$, and also refer to this  as the belief of the agent about the variable $s_{\tau:\tau'}$.

\section{Inference} \label{sec:inference}

\subsection{Action selection according to active inference} \label{sec:action-selection}

Let an agent be at time step $t$, having received observations $o_{1:t}$ and performed actions $a_{1:t-1}$. According to active inference an agent selects its next action by sampling a policy $\pi_t$ from the following distribution (equation (10) in \cite{dacosta2021active}): 
\begin{align} \label{eq:approx-post-policies-1}
   \sigma(- G(\pi_t| o_{1:t}, a_{1:t-1}) )
\end{align}
and selecting the action $a_t$ corresponding to that policy.\footnote{Note that at every time step a new policy is sampled, only the action $a_t$ for the corresponding time step $t$ is executed, and the rest of the actions in $\pi_t$ are discarded.} The function $\sigma$ denotes the softmax function defined in \eqref{eq:softmax-function} and $G$ is the expected free energy function given by
\begin{align}   \label{eq:efe}
    \begin{aligned}
        G(\pi_t| o_{1:t}, a_{1:t-1}) = - \bigg(&\bE_{q_t(o_{t+1:T}| \pi_t)}\bigg[\DKL \Big( q_t(s_{t+1:T} | o_{t+1:T}, \pi_t) \parallel q_t(s_{t+1:T} |\pi_t) \Big) \bigg] \\ 
        &+ \bE_{q_t(o_{t+1:T} | \pi_t)} \Big[\ln p_C(o_{t+1:T})\Big]\bigg).
    \end{aligned}
\end{align}
Note that according to \eqref{eq:approx-post-policies-1} the agent is more likely to sample policies $\pi_t$ that have a low expected free energy $G(\pi_t, o_{1:t}, a_{1:t-1})$. In Section \ref{sec:efe} we discuss equivalent formulations and different interpretations of the expected free energy. The distributions $q_t(o_{t+1:T}| \pi_t)$, $q_t(s_{t+1:T} | o_{t+1:T}, \pi_t)$, $q_t(s_{t+1:T}|\pi_t)$, needed for the calculation of $G$, are (approximate) posterior distributions of the generative model of the agent after having observed $o_{1:t}$ and performed $a_{1:t-1}$. In Section \ref{sec:state-inference} we describe how the agent infers these posterior distributions. The distribution $p_C$ is a preference distribution over observations that we assume is given to the agent. This distribution is distinct from the generative model $p$. 

\begin{remark}
    Note that in certain descriptions of active inference in discrete time also a variational free energy term $F$ appears in the distribution in equation \eqref{eq:approx-post-policies-1} (e.g. equation (B.9) in \cite{parr2022active}). This term is only relevant in specific cases that we will discuss in Remark \ref{remark:cartesian-product} in Appendix \ref{sec:vfe}. Furthermore a habit term $E$ is sometimes included that is also considered in Appendix \ref{sec:vfe}, but discarded here for simplicity. 
\end{remark}

\subsection{Expected free energy} \label{sec:efe}

Recall that equation \eqref{eq:efe} gives the following expression for the expected free energy function:
\begin{align*}  
    \begin{aligned}
        G(\pi_t| o_{1:t}, a_{1:t-1}) = - \bigg(&\bE_{q_t(o_{t+1:T}| \pi_t)}\bigg[\DKL \Big( q_t(s_{t+1:T} | o_{t+1:T}, \pi_t) \parallel q_t(s_{t+1:T} |\pi_t) \Big) \bigg] \\ 
        &+ \bE_{q_t(o_{t+1:T} | \pi_t)} \Big[\ln p_C(o_{t+1:T})\Big]\bigg).
    \end{aligned}
\end{align*}
The first term between the brackets on the RHS is called \emph{epistemic value} or \emph{information gain}. It measures the average change in belief about the future states $s_{t+1:T}$ due to receiving future observations $o_{t+1:T}$. The second term, known as \emph{utility}, quantifies the similarity between the expected future observation distribution and the preferred observation distribution. As previously mentioned, the agent is more likely to sample policies with low expected free energy, which correspond to high information gain and utility.

An equivalent formulation of expected free energy is given by
\begin{align} \label{eq:efe2}
    \begin{aligned}
        G(\pi_t| o_{1:t}, a_{1:t-1}) = \ &\bE_{q_t(s_{t+1:T} | \pi_t)}\Big[\mathrm{H}\big[p(o_{t+1:T} |s_{t+1:T})\big]\Big] \\ 
        + &\DKL\Big(q_t(o_{t+1:T} | \pi_t) \parallel p_C(o_{t+1:T})\Big) .
    \end{aligned}
\end{align}
The first term on the RHS is referred to as \emph{ambiguity}. It measures the average uncertainty an agent has about its future observations given knowledge of its future states. The second term is called \emph{expected complexity} or \emph{risk}. It represents the divergence between expected and preferred future observations. The agent favors policies with low ambiguity and risk. 

In Appendix \ref{further-details} we show that both expressions of the expected free energy are equal. 

In practice, often the following mean field approximations are made:
\begin{align}
    q_t(s_{t+1:T}) &= \prod_{\tau=t+1}^T q_t(s_\tau),\\
    p_C(o_{t+1:T}) &= \prod_{\tau=t+1}^T p_C(o_\tau).
\end{align}
Equation \eqref{eq:efe} and \eqref{eq:efe2} can then be written as follows:
\begin{align}  \label{eq:efe-mfa1}
    G(\pi_t| o_{1:t}, a_{1:t-1}) &= \sum_{\tau = t+1}^T G_\tau(\pi_t, o_{1:t}, a_{1:t-1}), 
\end{align}
with
\begin{align} \label{eq:efe-mfa2}
    G_\tau(\pi_t| o_{1:t}, a_{1:t-1}) &= \begin{aligned}[t]
        - \bigg( & \bE_{q_t(o_\tau| \pi_t)}\left[\DKL \Big( q_t(s_\tau | o_\tau, \pi_t) \parallel q_t(s_\tau|\pi_t) \Big) \right]  \\
    &+ \bE_{q_t(o_\tau | \pi_t)} \Big[\ln p_C(o_\tau)\Big]\bigg)
    \end{aligned} 
          \\
    &= \bE_{q_t(s_\tau | \pi_t)}\Big[\mathrm{H}\big[p(o_\tau |s_\tau)\big]\Big]  + \DKL\Big(q_t(o_\tau | \pi_t) \parallel p_C(o_\tau)\Big) .
\end{align}

\noeqref{eq:efe-mfa1, eq:efe-mfa2}
It is outside the scope of this paper to further derive or motivate the expected free energy. We refer the reader to Appendix B.2.5 in \cite{parr2022active} and \cite{da2024active,wei2024value,millidge2021whence} for more details.

\subsection{State inference} \label{sec:state-inference}

In this section we describe the simplest form of state inference, which is obtained by applying Bayes' rule. State inference methods as described in e.g.\ \cite{parr2022active,heins2022pymdp,smith2022step} can be thought of as computationally efficient approximations of what is described here. See Appendix \ref{sec:vfe} for more details on these methods. 

Above we defined $q_t$ to be the (approximate) posterior of the generative model given $o_{1:t}$, $a_{1:t-1}$. In this section we make the conditioning variables explicit and write $q( \, \cdot \, | o_{1:t}, a_{1:t-1})$ instead. 

\subsubsection*{Current and future state inference}

We start by studying the generative model that is assumed to be given to the agent. The generative model can be decomposed as follows: (see Figure \ref{fig:representation-genmod})
\begin{align}
    p(o_{1:T}, s_{1:T} | a_{1:T-1}) &= p(s_{1:T} | a_{1:T-1}) p(o_{1:T}| s_{1:T})\\
    p(s_{1:T} | a_{1:T-1}) &= p(s_1) \prod_{\tau=2}^{T} p(s_{\tau}| s_{\tau-1}, a_{\tau-1})\\
    p(o_{1:T}| s_{1:T}) &= \prod_{\tau=1}^{T} p(o_\tau| s_\tau).
\end{align}
At every time step the agent updates its belief about the current state it is in. Before having performed any observations, its belief about the current state is equal to the prior belief $p(s_1)$. After receiving $o_1$ it will update its belief using Bayes' rule:
\begin{align}
    q(s_1| o_1) \propto p(o_1|s_1) p(s_1). 
\end{align}
Subsequently, it will perform an action $a_1$ (selected as described in Section \ref{sec:action-selection}) and receive a next observation $o_2$. The belief about $s_2$ is given by 
\begin{align}
    q(s_2| o_{1:2}, a_1) &\propto p(o_2 | s_2, o_1, a_1) p(s_2| o_1, a_1)\\
    &= p(o_2|s_2) \sum_{s_1} p(s_2|s_1, o_1, a_1) p(s_1|a_1, o_1) \\
    &= p(o_2|s_2) \sum_{s_1} p(s_2|s_1, a_1) q(s_1| o_1). 
\end{align}
For a general $t$ the belief about $s_t$ is updated as follows:
\begin{align} \label{eq:update-state}
    q(s_t| o_{1:t},  a_{1:t-1}) \propto p(o_t|s_t) \sum_{s_{t-1}} p(s_t|s_{t-1}, a_{t-1}) q(s_{t-1}|o_{1:t-1}, a_{1:t-2}).  
\end{align}
For future time point $\tau > t$, the belief about the state  $s_\tau$ is given by
\begin{align} \label{eq:posterior-future-state}
    q(s_\tau| o_{1:t}, a_{1:\tau-1} ) = \sum_{s_{\tau-1}} p(s_\tau| s_{\tau-1}, a_{\tau-1})  q(s_{\tau-1}| o_{1:t}, a_{1:\tau-2}).
\end{align}
The belief about a future state given a future observation is calculated as follows:
\begin{align}  \label{eq:posterior-future-state-given-obs}
    q(s_\tau| o_\tau, o_{1:t},  a_{1:\tau-1} ) \propto p(o_\tau | s_\tau)  q(s_\tau| o_{1:t},  a_{1:\tau-1} ). 
\end{align}

\subsubsection*{Future observation inference}
In order to compute $G$ in \eqref{eq:efe}, we need a posterior distribution $q(o_\tau| o_{1:t}, a_{1:\tau-1})$ over future observations $o_\tau$. This can be computed as follows:
\begin{align} \label{eq:posterior-future-obs}
    q(o_\tau|o_{1:t}, a_{1:\tau-1}) = \sum_{s_\tau} p(o_\tau|s_\tau) q(s_\tau| o_{1:t},  a_{1:\tau-1} ).
\end{align}

\subsubsection*{Past, current, and future state inference}
In order to perform inference over states in the past, present and future (which is needed for the learning of the generative model and for the computation of the variational free energy over states), the agent can use the following formula:
\begin{align}
    q(s_{1:T}| o_{1:t}, a_{1:t-1}, \pi_t) &\propto p(o_{1:t}|s_{1:T}, a_{1:t-1}, \pi_t) p(s_{1:T} | a_{1:t-1}, \pi_t) \\
    &=  p(o_{1:t}|s_{1:t}) p(s_{1:T} | a_{1:t-1}, \pi_t) \\
    &= \prod_{\tau=1}^t p(o_\tau | s_\tau) \ p(s_1) \prod_{\tau=2}^T p(s_\tau|s_{\tau-1}, a_{\tau-1}), \label{eq:posterior-states-past-future}
\end{align}
where we use $a_{\tau-1}$ in the last term for elements of both $a_{1:t-1}$  and $\pi_t$.

\section{Learning} \label{sec:learning}

In the above section, we have assumed that the agent has access to a generative model $p(o_{1:T}, s_{1:T} | a_{1:T-1}, \theta)$. In this section we discuss how the parameter $\theta$ of this model is learned. The generative model consists of three (conditional) categorical distributions that are parametrized by $\theta = (\theta^D, \theta^A, \theta^B)$ in the following way:
\begin{align}
    &p(s_1^{(j)}| \theta^D) = \theta^D_j, \label{gen-mod1} \\
    &p(o_\tau^{(i)}| s_\tau^{(j)}, \theta^A) = \theta^A_{ij}, \label{gen-mod2} \\
    &p(s_{\tau}^{(j)}| s_{\tau-1}^{(k)}, a_{\tau-1}^{(l)}, \theta^B) = \theta^B_{jkl}, \label{gen-mod3}
\end{align}
where we have enumerated the elements of the observation, action and state space with the bracketed superscript $^{(\cdot)}$. 
In order to learn the parameters of the generative model, we adopt a Bayesian belief updating scheme with a Dirichlet prior. (See Appendix \ref{sec:ap-learning} for details on this.)  More specifically, the prior over $\theta$ is parametrized by $\alpha = (\alpha^D, \alpha^A, \alpha^B)$ and is given by 
\begin{align}
    p(\theta|\alpha) &= p(\theta^D | \alpha^D) \prod_j p(\theta^A_{\bullet j} | \alpha^A) \prod_{k,l} p(\theta^B_{\bullet kl} | \alpha^B) \label{gen-mod-prior1}\\
    p(\theta^D | \alpha^D) &\propto  \prod_j \left(\theta^D_j\right)^{\alpha^D_j - 1}, \label{gen-mod-prior2} \\
    p(\theta^A_{\bullet j} | \alpha^A) &\propto  \prod_i \left(\theta^A_{ij}\right)^{\alpha^A_{i j} - 1}, \label{gen-mod-prior3}\\
    p(\theta^B_{\bullet kl} | \alpha^B) &\propto  \prod_j \left(\theta^B_{jkl}\right)^{\alpha^B_{jkl} - 1}, \label{gen-mod-prior4}
\end{align}
where $\theta_{\bullet j}$ denotes the vector $(\theta_{1j},\ldots, \theta_{nj})$.

Now after performing actions $a_{1:T-1}$ and receiving observations $o_{1:T}$ we want to update our belief about $\theta$ according to Bayes' rule. The true posteriors over these parameters are not Dirichlet distributions. (See Appendix \ref{sec:ap-learning} for details.) The active inference literature suggests to set the approximate posterior distribution to be a Dirichlet distribution and update the hyperparameter $\alpha$ in the following way (equation (B.12) in \cite{parr2022active} and equation (21), (A.6) and (A.7) in \cite{dacosta2021active}):
\begin{align}
    {\alpha^D_j}'  &= \alpha^D_j + q_T\left(s_1^{(j)}\right), \label{eq:approx-update-D}\\
    {\alpha^A_{ij}}'  &= \alpha^A_{ij} + \sum_{\tau=1}^T \mathbbm{1}_{o^{(i)}}(o_\tau) q_T\left(s_\tau^{(j)}\right), \label{eq:approx-update-A}\\
    {\alpha^B_{jkl}}'  &= \alpha^B_{jkl} + \sum_{\tau=2}^{T} q_T\left(s_{\tau}^{(j)}\right) q_T\left(s_{\tau-1}^{(k)} \right)  \mathbbm{1}_{a^{(l)}}(a_{\tau-1}). \label{eq:approx-update-B}
\end{align}
The distributions $q_T\left(s_\tau\right), \tau \in \{1,\ldots,T\}$ are approximate posteriors obtained\footnote{See Section \ref{sec:state-inference}.} using the current version of the generative model (before $\theta$ has been updated).  
In Appendix \ref{sec:ap-learning} we elaborate on the origin of this learning rule. 

Note the similarity with the standard update rule for Dirichlet priors given in \eqref{eq:true-update-dirichlet}. In the standard update rule the element $\alpha_{i^*}$ of the hyperparameter corresponding to the observation $x^{(i^*)}$ is incremented by 1, which makes this observation more likely in the updated distribution. In the updates \eqref{eq:approx-update-D}--\eqref{eq:approx-update-B} \noeqref{eq:approx-update-A} the hyperparameters are incremented by the amount of posterior belief in that state or state transition, e.g.\ $\alpha^D_j$ is incremented by $q\left(s_1^{(j)}| o_{1:T}, a_{1:T-1} \right)$. 

In  order to go from a Dirichlet distribution $p(\theta | \alpha)$ to an actual value of the parameter that can be used for the generative model, the mean of the distribution can be used, which is given by
\begin{align}
    \hat{\theta}_i &= \bE_{p(\theta | \alpha)}[\theta_i] \\
    &= \frac{\alpha_i}{\sum_j \alpha_j}. \label{eq:update-theta}
\end{align}
\noeqref{eq:update-theta}For example, after the learning step, the new distribution over $s_1$ is given by 
\begin{align}
    p(s_1^{(j)}| \hat\theta^D) = \frac{{\alpha^D_j}'}{\sum_j {\alpha^D_k}'}.
\end{align}

This concludes the discussion of learning in the context of the active inference framework.

 \section{Example: T-maze example}

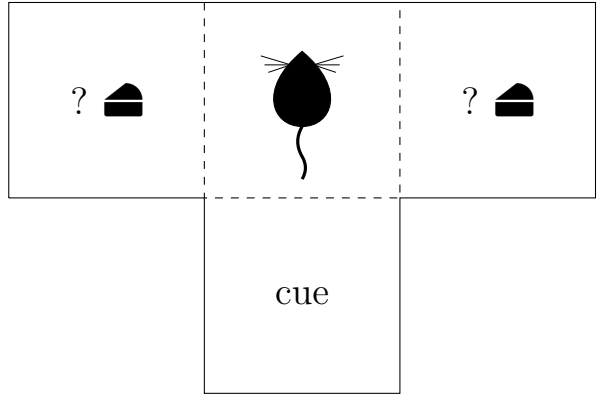
\begin{figure}
    \centering
\begin{tikzpicture}
    \pgfmathsetmacro{\gs}{2.6}
    \draw (0,0) -- (\gs, 0) -- (\gs, -\gs) -- (2*\gs, -\gs) -- (2*\gs, 0) -- (3*\gs, 0) -- (3*\gs, \gs) -- (0, \gs) -- (0,0);
    \draw [dashed] (\gs, \gs) -- (\gs, 0) -- (2*\gs, 0) -- (2*\gs, \gs);

    \pgfmathsetmacro{\xc}{1.5*\gs}
    \pgfmathsetmacro{\yc}{1.8}

    \coordinate (a) at (1.5*\gs, 1.95); 

    \draw [fill=black] (a) .. controls ($ (a) + (-.6 ,-.5) $) and ($ (a) + (-.4 , -1) $) .. ($ (a) + (0 ,-1) $) .. controls ($ (a) + (.4 , -1) $) and ($ (a) + (.6 ,-.5) $) .. cycle;

    \coordinate (b) at ($(a) + (-.15, -.18)$);

    \draw [very thin] (b) -- ($ (b) + (-.4, .12) $); 
    \draw [very thin] (b) -- ($ (b) + (-.35, 0) $); 
    \draw [very thin] (b) -- ($ (b) + (-.3, -.1) $); 

    \coordinate (c) at ($(a) + (.15, -.18)$);

    \draw [very thin] (c) -- ($ (c) + (.4, .12) $); 
    \draw [very thin] (c) -- ($ (c) + (.35, 0) $); 
    \draw [very thin] (c) -- ($ (c) + (.3, -.1) $); 

    \draw [very thick] ($(a) + (0, -1)$) to [bend right] ($(a) + (0,-1.4)$) to [bend left] ($(a) + (0,-1.7)$);

    \node at (.5*\gs, .5*\gs) {\Large{? \faCheese}};
    \node at (2.5*\gs, .5*\gs) {\Large{? \faCheese}};
    \node at (1.5*\gs, -.5*\gs) {\Large cue};

\end{tikzpicture}
\caption{T-maze environment}
\label{fig:tmaze}
\end{figure}

In this section we discuss the action selection mechanism of an active inference agent in the T-maze environment depicted in Figure \ref{fig:tmaze}. (See also Section 7.3 of \cite{parr2022active} and \cite{heins2021tmaze}.) This illustrates the theory of action selection and state inference that is presented in Section \ref{sec:inference}.

\subsection*{Description of the (internal) generative model of the agent}

\subsubsection*{State, observation and action spaces}

The internal state space of the agent\footnote{In this example the state of the environment (generative process) is the same as the state space of the internal world model of the agent (generative model). Note that this is in general not the case. In a more realistic setting the state space of the environment will be much more complex than the internal state space. } $\cS$ has two dimensions,\footnote{The dimensions of the state space are  sometimes referred to as state factors.}, \emph{Location} and \emph{Reward condition}, and can be described as follows:
\begin{align}
    \cS &= \cS^L \times \cS^R, \\
    \cS^L &= \{ \texttt{center}, \ \texttt{right arm}, \ \texttt{left arm}, \texttt{cue location} \}, \\
   \cS^R &= \{ \texttt{reward on right}, \ \texttt{reward on left} \}.
\end{align}
A typical element of the state space is written as $s = (s^L, s^R)$. 

The observation space $\cO$ has three dimensions\footnote{The dimensions of the observation space are sometimes referred to as observation modalities.}\footnote{Note that we follow here the description from \cite{heins2021tmaze}. In \cite{parr2022active} the cue observation is absorbed into the location observation.}, \emph{Location}, \emph{Reward}, and \emph{Cue}, and can be described as follows:
\begin{align}
    \cO &= \cO^L \times \cO^R \times \cO^C, \\
    \cO^L &= \{ \texttt{center}, \ \texttt{right arm}, \ \texttt{left arm}, \ \texttt{cue location} \}, \\
    \cO^R &= \{ \texttt{no reward}, \  \texttt{reward}, \ \texttt{loss} \},\\
    \cO^C &= \{ \texttt{cue right}, \ \texttt{cue left} \} .
\end{align}
A typical element of the observation space is written as $o = (o^L, o^R, o^C)$. 

The space of actions $\cA$ is described by 
\begin{align}
    \begin{aligned}
        \cA = \{ & \texttt{move to center}, \ \texttt{move to right arm}, \\
        & \texttt{move to left arm}, \ \texttt{move to cue location} \}.
    \end{aligned}
\end{align}
Note that these actions are always available, independent of the current location of the agent. A typical element of the action space is written as $a$. 

In the following we use \texttt{[dir]} as a placeholder for \texttt{left} and \texttt{right}, and \texttt{[loc]} as a placeholder for the four locations \texttt{center}, \texttt{right arm}, \texttt{left arm} and \texttt{cue location}. 

\subsubsection*{Observation kernel $p(o|s)$}

We now specify the observation kernel $p(o|s)$ of the generative model of the agent. First note that the observation dimensions are independent, that is
\begin{align}
    p(o|s) = p(o^L|s) p(o^R|s).
\end{align}
The beliefs about the location observation given the state are modelled as follows:
\begin{align}
    p(o^L|s)  = \begin{cases}
        1 \quad \text{if} \  o^L = s^L \\
        0 \quad \text{otherwise}
    \end{cases},
\end{align}
which implies that the location can be unambiguously inferred from the observation. 

The beliefs about the reward observation given the state are modelled as follows:
\begin{align}
    p(o^R = \texttt{no reward} | s^L \in \{\texttt{center}, \texttt{cue location} \}, s^R) &= 1, \\ 
    p(o^R = \texttt{no reward} | s^L \notin \{\texttt{center}, \texttt{cue location} \}, s^R) &= 0, \\ 
    p(o^R = \texttt{reward} | s^L = \texttt{[dir] arm}, s^R = \texttt{reward on [dir]}) &=  0.98 , \\ 
    p(o^R = \texttt{loss} | s^L = \texttt{right arm}, s^R = \texttt{reward on left} ) &= 0.98, \\
    p(o^R = \texttt{loss} | s^L = \texttt{left arm}, s^R = \texttt{reward on right} ) &= 0.98.
\end{align}
This implies that the agent observes \texttt{no reward} when it is in location \texttt{center} or \texttt{cue location}, it observes \texttt{reward} when it is in the same arm as specified by the reward condition with high probability, and it observes \texttt{loss} when it is in the opposite arm of the reward condition with high probability. 

The beliefs about the cue observation given the state are modelled as follows:
\begin{align}
    &p(o^C = \texttt{cue [dir]} |s^L = \texttt{cue location}, s^R = \texttt{reward on [dir]}) = 1, \\
    &p(o^C = \texttt{cue [dir]} | s^L \in \cS^L \setminus \{\texttt{cue location}\}, s^R ) = 0.5 .
\end{align}
This implies that the cue observation is completely informative about the reward condition when the agent is at the cue location, and otherwise independent of the actual reward condition. 

\subsubsection*{Transition dynamics kernel  $p(s_{\tau+1}|s_\tau, a_\tau)$}

We continue by describing the  transition dynamics kernel $p(s_{\tau+1}|s_\tau, a_\tau)$.
\begin{align}
    p(s^L_{\tau+1}&=\texttt{[loc]} | s^L_{\tau} \in \{\texttt{center},  \texttt{cue location} \}, s^R_\tau, a_\tau = \texttt{go to [loc]} ) = 1,\\
    p(s^L_{\tau+1}&=\texttt{[dir] arm} | s^L_{\tau} = \texttt{[dir] arm}, s^R_\tau, a_\tau ) = 1.
\end{align}
This implies that if the agent is in \texttt{center} or \texttt{cue location} it will be in the location specified by the action in the next time step. If it is in one of the arms however, it will stay there, independent of the choice of action $a_\tau$. 
\begin{align}
    p(s^R_{\tau+1} = \texttt{reward on [dir]} | s^R_{\tau} = \texttt{reward on [dir]}, s^L_\tau, a_\tau) = 1.
\end{align}
This implies that the reward condition stays constant throughout the trajectory. 

\subsubsection*{Preference distribution $p_C$ and prior over states $p_D$}
The unnormalized preference distribution $p_C$ can be chosen to favor observations with \texttt{reward} and discourage observations with \texttt{loss} as follows:
\begin{align}
    &p_C((\texttt{[loc]}, \texttt{no reward}, \texttt{cue [dir]} )) = 2, \\
    &p_C((\texttt{[loc]}, \texttt{reward}, \texttt{cue [dir]} )) = 3, \\
    &p_C((\texttt{[loc]}, \texttt{loss}, \texttt{cue [dir]} )) = 1. 
\end{align}

Finally we let the prior belief over states $p_D$ be uniform. 

\subsection*{Action selection procedure}

We will now simulate the trajectory of an agent acting according to active inference. We set the time horizon to $T=3$, which implies that the policies will have length 2. 

\subsubsection*{Time step 1}

The agent starts by receiving an observation $o_1$ = (\texttt{center}, \texttt{no reward}, \texttt{cue right}). It now updates its beliefs about the current state such that 
\begin{align}
    q_t(s_1^L = \texttt{center}, s_1^R = \texttt{reward on [dir]}) = 0.5.
\end{align} 
Subsequently it computes its beliefs about future states and observations given a policy using equation \eqref{eq:posterior-future-state}, \eqref{eq:posterior-future-state-given-obs}, \eqref{eq:posterior-future-obs}. For example for $\pi_1^*$ = (\texttt{move to cue location}, \texttt{move to left arm}) we have
\begin{align}
    q_{1}(s_2^L = \texttt{cue location}, s_2^R = \texttt{reward on [dir]} \mid \pi_1 = \pi_1^*) = 0.5,  \label{eq:qs2unconditioned}\\
    q_{1}(s_3^L = \texttt{left arm}, s_3^R = \texttt{reward on [dir]} \mid \pi_1 = \pi_1^*) = 0.5,
\end{align}
and
\begin{align}
    q_{1}(o_2^L = \texttt{cue location}, o_2^R = \texttt{no reward}, o_2^C = \texttt{{cue [dir]}} \mid \pi_1 = \pi_1^*) = 0.5, \\
    q_{1}(o_3^L = \texttt{left arm}, o_3^R = \texttt{reward}, o_3^C = \texttt{{cue [dir]}} \mid \pi_1 = \pi_1^*) = 0.25, \\
    q_{1}(o_3^L = \texttt{left arm}, o_3^R = \texttt{loss}, o_3^C = \texttt{{cue [dir]}} \mid \pi_1 = \pi_1^*) = 0.25,
\end{align}
and for example for $o_2^*$ = (\texttt{cue location}, \texttt{no reward}, \texttt{cue left}) we have
\begin{align}
    q_{1}(s_2^L = \texttt{cue location}, s_2^R = \texttt{reward on left}  \mid o_2=o_2^*, \pi_1 = \pi_1^*) = 1. \label{eq:qs2conditioned}
\end{align}
Note here the reduction of uncertainty about $s_2$ due to the observation $o_2^*$, represented by the epistemic value defined in Section \ref{sec:efe} given by the KL divergence between the distributions \eqref{eq:qs2conditioned} and \eqref{eq:qs2unconditioned}. 

The agent now computes $G$ and plugs this into equation \eqref{eq:approx-post-policies-1} and gets the following posterior distribution over policies: 

\begin{center}
    \begin{tabular}{ l | c c c c}
        \begin{tabular}{@{}l@{}} \multicolumn{1}{r}{ \ \ \ \ \ \ \ \ $a_2$  $\rightarrow$} \\ $a_1$  $\downarrow$ \end{tabular} & \texttt{center} & \texttt{right arm} & \texttt{left arm} & \texttt{cue location} \\ 
     \hline
     \texttt{center}       & 0.022 & 0.041 & 0.041 & 0.046\\  
     \texttt{right arm }   & 0.041 & 0.075 & 0.075 & 0.083\\  
     \texttt{left arm}     & 0.041 & 0.075 & 0.075 & 0.083 \\
     \texttt{cue location} & 0.046 & 0.083 & 0.083 & 0.091

    \end{tabular}
    \end{center}

and in this scenario it samples a policy with as first action \texttt{move to cue location} with highest probability. 

\subsubsection*{Time step 2}

After having performed action $a^*_1 =$ \texttt{move to cue location}, the next observation it receives is $o^*_2$ = (\texttt{cue location}, \texttt{no reward}, \texttt{cue right}). Its belief about the current state is now given by
\begin{align}
    q_{2}(s_2^L = \texttt{cue location}, s_2^R = \texttt{reward on right}) = 1.
\end{align} 
For instance, when $\pi_2^*$ = (\texttt{move to left arm}), the beliefs about future states are
\begin{align}
    q_{2}(s_3^L = \texttt{left arm}, s_3^R = \texttt{reward on right} \mid \pi_2 = \pi_2^*) = 1,
\end{align}
and the observation beliefs
\begin{align}
    q_{2}(o_3^L = \texttt{left arm}, o_3^R = \texttt{reward}, o_3^C = \texttt{{cue [dir]}} \mid \pi_2 = \pi_2^*) = 0.01, \\
    q_{2}(o_3^L = \texttt{left arm}, o_3^R = \texttt{loss}, o_3^C = \texttt{{cue [dir]}} \mid \pi_2 = \pi_2^*) = 0.49 .
\end{align}
Since all uncertainty has already been taken away by the last observation, conditioning on $o_3^*$ = (\texttt{left arm}, \texttt{reward}, \texttt{cue left}) will make no difference to the belief about the state $s_3$, i.e.
\begin{multline}
    q_{2}(s_3^L = \texttt{left arm}, s_3^R = \texttt{reward on right} \mid o_3 = o_3^*, \pi_2 = \pi_2^*) = \\  q_{2}(s_3^L = \texttt{left arm}, s_3^R = \texttt{reward on right} \mid \pi_2 = \pi_2^*) = 1,
\end{multline}
which will cause the epistemic value term in $G$ to be zero. 

The agent now calculates $G$ again and obtains the following distribution over policies: 

\begin{center}
    \begin{tabular}{ l | c c c c}
        $a_2$  & \texttt{center} & \texttt{right arm} & \texttt{left arm} & \texttt{cue location} \\ 
        \hline
           & 0.20 & 0.52 & 0.08 & 0.20
    \end{tabular},
\end{center}

and will then sample the policy (\texttt{move to right arm}) with highest probability. 

\section*{Acknowledgements}   

The authors would like to thank Thomas Parr, Conor Heins, Ryan Smith, Beren Millidge, Pablo Lanillos, Sean Tull, Stephen Mann, Pradeep Kumar Banerjee, Frank Röder and Lance Da Costa for helpful discussions and comments and acknowledge the support of the Deutsche Forschungsgemeinschaft Priority Programme “The Active Self” (SPP 2134).

\printbibliography

\newpage

\appendix

\section{Variational free energy minimization}
\label{sec:vfe}

\subsection*{Preliminaries}

In this section we use $x$ for a general observed variable and $z$ for a general latent variable. 

Let $p(x,z) = p(x|z)p(z)$ be a generative model. In order to perform inference over the latent variables after making an observation $x$, one has to compute the posterior $p(z|x)$. This posterior is often hard to compute directly. One can instead approximate this posterior by finding the distribution $q_x(z)$ in a family of distributions $\cQ$ that minimizes the following function:
\begin{align}
    F(\tilde{q}|x) = \sum_z \tilde{q}(z) \left(\ln \tilde{q}(z) - \ln p(x, z) \right),
\end{align}
called the variational free energy. Note that we distinguish notationally $\tilde{q}$, which is a generic element of $\cQ$ and a variable in $F$, and $q_x$, which is the minimizer of $F$ for a fixed $x$, i.e.
\begin{align}
    q_x = \argmin_{\tilde{q} \in \cQ} F(\tilde{q}| x).
\end{align}
Note that when $\cQ$ is large enough, for example when $z$ is discrete and $\cQ$ is the set of all probability distributions over $z$, then the minimizer of the free energy is equal to the exact posterior distribution and we have
\begin{align}
    q_x(z) &= p(z|x).
\end{align}
We can replace $\ln p(x,z)$ in $F$ by a general function $f$, i.e.
\begin{align}
    F_f(\tilde{q}|x) = \sum_z \tilde{q}(z) \left(\ln \tilde{q}(z) - f(x, z) \right).
\end{align}
The minimizer can be found by substituting $g(x,z) = e^{f(x,z)}$ as follows: 
\begin{align}
    F_f(\tilde{q},x) &= \sum_z \tilde{q}(z) \left(\ln \tilde{q}(z) - \ln g(x,z)\right) \\
    &=  \left(\sum_z \tilde{q}(z) \left(\ln \tilde{q}(z) - \ln \frac{g(x,z)}{\sum_{z'} g(x,z')} \right) \right)  + \ln \sum_{z'} g(x,z').
\end{align}
If $\cQ$ is again large enough, the minimizer is given by: 
\begin{align}
    q_x(z) &=  \frac{g(x,z)}{\sum_z g(x,z)} \\
    &=  \frac{e^{f(x,z)}}{\sum_z e^{f(x,z)}} \\ 
    &= \sigma(f(x,z)), \label{eq:exactSolution}
\end{align}
where $\sigma$ is the softmax function defined in \eqref{eq:softmax-function} and in the case that $f(x,z) = \ln p(x,z)$ we have
\begin{align}
    q_x(z) &= \sigma(\ln p(x,z))\\
    &= p(z|x).
\end{align}

The variational free energy can be written as follows:
\begin{align}
    F(\tilde{q}|x) &= \DKL(\tilde{q}(z) \parallel p(z|x)) - \ln p(x) \\
    & \geq - \ln p(x),
\end{align}
which shows that the negative of the variational free energy is a lower bound on the evidence (ELBO). By minimizing $F$ w.r.t.\ $\tilde{q}$ we get the following approximate equality (equation (B.2) in \cite{parr2022active}):
\begin{align} \label{eq:approximateEvidence}
    F(q_x| x) \approx -\ln p(x).
\end{align}

\subsection*{Variational free energy minimization in active inference} 

Active inference adopts the perspective that perception, action selection and learning can be interpreted as minimizing one single variational free energy function $F$. In its complete form it can be written as follows:
\begin{align} \label{eq:vfe-un}
    F(\tilde{q}| o_{1:t}, a_{1:t-1}) = \bE_{\tilde{q}(s_{1:T}, \theta, \pi_t)} \left[  
        \ln \tilde{q}(s_{1:T}, \theta, \pi_t) - \ln p(s_{1:T}, o_{1:t}, \theta, \pi_t|a_{1:t-1})
    \right].
\end{align}
The distributions in the family $\cQ$ are assumed to factorize as follows:
\begin{align}
    q(s_{1:T}, \theta, \pi) = q(\theta^D)q(\theta^A)q(\theta^B) q(\pi) \prod_{\tau=1}^T q(s_\tau), 
\end{align}
which is  sometimes referred to as the mean-field approximation. Now $F$ can be written as follows: 
\begin{multline} \label{eq:vfe-complete}
    F(\tilde{q}| o_{1:t}, a_{1:t-1}) = \bE_{\tilde{q}(s_{1:T}, \theta, \pi_t)} \bigg[ 
        \ln \tilde{q}(\pi_t) + \ln \tilde{q}(\theta) + \sum_{\tau=1}^T \ln \tilde{q}(s_\tau) -
         \ln p(\pi_t) \\ 
         - \ln p(\theta) - \ln p(s_1 | \theta^D) -  
        \sum_{\tau=1}^t \ln p(o_\tau | s_\tau , \theta^A) - \sum_{\tau=2}^T \ln  p(s_\tau| s_{\tau-1}, a_{\tau-1}, \theta^B)
    \bigg],
\end{multline}
where we use $a_{\tau-1}$ in the last term for elements of both $a_{1:t-1}$  and $\pi_t$. 

\subsubsection*{Perception}

For studying perception, equation \eqref{eq:vfe-complete} can be rewritten as follows: 
\begin{align}
    F(\tilde{q}| o_{1:t}, a_{1:t-1}) = \bE_{\tilde{q}(\pi_t)}  \big[ 
        F_{\pi_t}(\tilde{q}| o_{1:t}, a_{1:t-1})
    \big]   + C_{\setminus s}
\end{align}
where $C_{\setminus s}$ is independent of $\tilde{q}(s_{1:T})$ and 
\begin{multline} \label{eq:vfe-fixed-policy}
    F_{\pi_t}(\tilde{q}| o_{1:t}, a_{1:t-1}) = \\
    \bE_{\tilde{q}(s_{1:T})} \left[
        \ln \tilde{q}(s_{1:T}) - \bE_{\tilde{q}(\theta)}[ \ln p(s_{1:T}, o_{1:t} | \theta, a_{1:t-1}, \pi_t) ]\right] .
\end{multline}
Perception according to active inference is minimizing $F_{\pi_t}$ w.r.t.\ $\tilde{q}(s_{1:T})$. The minimizer is written $q_t(s_{1:T}| \pi_t) = q(s_{1:T}| \pi_t, o_{1:t}, a_{1:t-1})$.

We can use the factorizing properties to rewrite \eqref{eq:vfe-fixed-policy} as follows:
\begin{multline} 
    F_{\pi_t}(\tilde{q}| o_{1:t}, a_{1:t-1}) = \sum_{\tau=1}^T \bE_{\tilde{q}(s_{\tau})} [         \ln \tilde{q}(s_{\tau}) ] - 
    \bE_{\tilde{q}(s_1, \theta^D)}[ \ln p(s_1 | \theta^D) ]  \\ 
    - \sum_{\tau=1}^t \bE_{\tilde{q}(s_\tau, \theta^A)} p(o_\tau | s_\tau , \theta^A) - \sum_{\tau=2}^T \bE_{\tilde{q}(s_\tau, s_{\tau-1}, \theta^B)} p(s_\tau | s_{\tau-1} , a_\tau, \theta^B),
\end{multline}
which is equivalent to equation (6) in \cite{dacosta2021active}.\footnote{Note that in \cite{dacosta2021active} only the parameter $\theta^A$ is treated as a variable and $\theta^D$ and $\theta^B$ are considered fixed.} One can also get rid of the expectations over $\theta$ by replacing them by an estimator $\hat{\theta}$. Then we get 
\begin{multline} 
    F_{\pi_t}(\tilde{q}| o_{1:t}, a_{1:t-1}) = \sum_{\tau=1}^T \bE_{\tilde{q}(s_{\tau})} [         \ln \tilde{q}(s_{\tau}) ] - 
    \bE_{\tilde{q}(s_1)}[ \ln p(s_1 | \hat{\theta}^D) ]  \\ 
    - \sum_{\tau=1}^t \bE_{\tilde{q}(s_\tau)} p(o_\tau | s_\tau , \hat{\theta}^A) - \sum_{\tau=2}^T \bE_{\tilde{q}(s_\tau, s_{\tau-1} )} p(s_\tau | s_{\tau-1} ,a_\tau, \hat{\theta}^B), 
\end{multline}
which is equivalent to (B.4) in \cite{parr2022active}.

\subsubsection*{Learning}

Updating the $\theta$ parameter (learning) happens at the end of an episode ($t=T$). The agent has observed $o_{1:T}$ and performed $a_{1:T-1}$. The variational free energy from equation \eqref{eq:vfe-complete} can be written as follows:  
\begin{multline} \label{eq:vfe-learning-fixed-qs1}
    F(\tilde{q}| o_{1:T}, a_{1:T-1}) = \\
    \bE_{\tilde{q}(\theta)} \left[  
        \ln \tilde{q}(\theta) - \bE_{q_T(s_{1:T})} [\ln p(o_{1:T}, s_{1:T}, \theta| a_{1:T-1})]  \right] + C_{\setminus \theta},
\end{multline}
where $C_{\setminus \theta}$ is independent of $\tilde{q}(\theta)$ and we have fixed $q_T(s_{1:T})$ to be the approximate posterior over states, inferred using the current (not-updated) belief $p(\theta)$. If the current belief $p(\theta)$ is a Dirichlet distribution with hyperparameter $\alpha$, then the minimizer $q_T(\theta)$ of $F$ will also be a Dirichlet distribution with hyperparameter $\alpha'$ as given in \eqref{eq:approx-update-D}--\eqref{eq:approx-update-B}. In Appendix \ref{sec:ap-learning} we derive this, and relate it to a well known variational inference algorithm called coordinate ascent variational inference (CAVI).

\subsubsection*{Action selection}

Finally we can also view action selection as the minimization of the variational free energy function \eqref{eq:vfe-complete}. We can rewrite this function as follows:
\begin{align}
    F(\tilde{q}| o_{1:t}) &= \bE_{\tilde{q}(\pi)} \left[ \ln \tilde{q}(\pi) - \ln p(\pi) + \bE_{\tilde{q}(s_{1:T})} \left[ \ln \tilde{q}(s_{1:T}) - \ln p(o_{1:t}, s_{1:T} | \pi ) \right] \right] + C_{\setminus \pi} \\
    &= \bE_{\tilde{q}(\pi)} \left[ \ln \tilde{q}(\pi) - \ln p(\pi) + F_\pi(\tilde{q}(s_{1:T})| o_{1:t}) \right] + C_{\setminus \pi}, \label{eq:vfe-policies}
\end{align}
where $F_\pi$ is defined in \eqref{eq:vfe-fixed-policy}, $C_{\setminus \pi}$ is independent of $\tilde{q}(\pi)$, and we replaced the expectation over $\tilde{q}(\theta)$ by an estimator $\hat{\theta}$ and suppress the dependence of the generative model $p$ on $\hat{\theta}$ in the notation. (This is equivalent to the second line in equation (B.7) in \cite{parr2022active}.) What is important to note here, is that the agent is trying to infer an action sequence (policy) $\pi$ of both future and past actions. We have therefore dropped the dependence on $a_{1:t-1}$ in both $F$ and $F_\pi$, and instead $\pi$ is a sequence of action starting at $\tau=1$ instead of $\tau = t$. In Section \ref{sec:inference}, we always fixed the past actions to the actions that were actually performed, which is no longer the case here. 

We minimize $F$ w.r.t.\ $\tilde{q}(\pi)$ and $\tilde{q}(s_{1:T})$ and using \eqref{eq:exactSolution} we get for the minimizers respectively 
\begin{align} \label{eq:posterior-policies}
    q_t(\pi) = \sigma \left( - \ln p(\pi) + F_\pi(q_t(s_{1:T})| o_{1:t}) \right),
\end{align}
and $q_t(s_{1:T})$ is the minimizer of $F_\pi$. Now we can use 
\begin{align}
    p(\pi) = \sigma( \ln E(\pi) - G(\pi_t| o_{1:t}, a_{1:t-1})), 
\end{align}
which corresponds to the last line equation (B.7) in \cite{parr2022active}.\footnote{It can be argued that calling this a prior is incorrect, since it actually depends on the observations $o_{1:t}$.}
The term $E(\pi)$ is a habit term, signifying what policies the agent is usually exercising. Plugging this back into \eqref{eq:posterior-policies} gives 
\begin{align} \label{eq:approx-post-policies-2}
    q_t(\pi) &= \sigma \left( - \ln E(\pi) + G(\pi_t| o_{1:t}, a_{1:t-1}) + F_\pi(q_t| o_{1:t}) \right),
\end{align}
which corresponds to equation (B.9) in \cite{parr2022active}. 

\begin{remark} \label{remark:cartesian-product}
    We now try to interpret this derivation conceptually. Note that due to \eqref{eq:approximateEvidence} we have the following approximate equality: 
    \begin{align}
        F(\tilde{q}| o_{1:t}) &= \bE_{\tilde{q}(\pi)} \left[ \ln \tilde{q}(\pi) - \ln p(\pi) + F_\pi(\tilde{q}(s_{1:T}), o_{1:t}) \right] + C_{\setminus \pi} \\
        &\approx \bE_{\tilde{q}(\pi)} \left[ \ln \tilde{q}(\pi) - \ln p(\pi) - \ln p(o_{1:t} | \pi) \right] + C_{\setminus \pi} \\
        &= \bE_{\tilde{q}(\pi)} \left[ \ln \tilde{q}(\pi) -\ln p(o_{1:t}, \pi) \right] + C_{\setminus \pi},
    \end{align}
    which implies that the minimizer $q_t(\pi)$ is approximately equal to the posterior $p(\pi | o_{1:t})$. In other words, this says that we select the policy that is most probable given the past observations. That is, the agent forgets which past actions it has performed, and tries to infer these based on the past observations. Then it tries to find the most likely sequence of future actions to go with this sequence of past actions. Selecting future actions in this way however only makes sense when certain past action sequences make certain future action sequences more likely. For example, let our action space consist of two actions $\{ \texttt{left}, \texttt{right} \}$ and policies consist of sequences of actions of length two. Now suppose that the prior over policies dictates that the agent almost certainly performs the policies $(\texttt{left}, \texttt{left})$ or $(\texttt{right}, \texttt{right})$. This implies that having inferred the first action gives the agent more information about the most likely next action. However, if both next actions are equally likely given a first action, according to the prior, then the likelihood term $p(o_{1:t}|\pi)$ does not have any information about the next action. Note that in the calculation of $G$ the past actions are fixed to the actions that have been performed. Therefore $G$ will not make certain future action sequences more likely based on possible past action sequences. Therefore, the term $F_\pi$ in \eqref{eq:approx-post-policies-2} only becomes relevant when the habit term $E$ makes certain future action sequences more likely based on past action sequences. 
\end{remark}

\section{Learning Preliminaries} \label{sec:ap-learning}

\subsection*{Bayesian belief updating}

The learning process of an active inference agent is formulated as Bayesian belief updating over the parameters. In general, Bayesian belief updating can be described as follows. Let $\theta$ be the parameter of a model $p_\theta$ we want to learn and $x$ the output of this model. We start with a prior belief $p(\theta | \alpha)$ which is parametrized by the hyperparameter $\alpha$. Now our posterior belief about $\theta$ is given by the distribution $p(\theta |x, \alpha)$ which is obtained by Bayes' rule. In some special cases\footnote{For details, see the theory of conjugate priors.}, the posterior distribution belongs to the same parametrized family as the prior, such that $p(\theta| x , \alpha) = p(\theta | \alpha')$. Then the learning can be summarized by the update from $\alpha$ to $\alpha'$. 

\subsection*{Categorical model without latent variables}

Now we let the model be a categorical distribution over elements $\{x^{(1)}, \ldots, x^{(n)}\}$ parametrized by $\theta = (\theta_1, \ldots, \theta_n)$. That is,
\begin{align}
    p(x^{(i)}|\theta) = \theta_i, \ \ \  \forall i \in \{1,\ldots,n\}. 
\end{align}
The prior over the parameter $\theta$ is given by the Dirichlet distribution parametrized by the hyperparameter $\alpha = (\alpha_1, \ldots, \alpha_n)$. That is, 
\begin{align}
    p(\theta| \alpha) \propto \prod_i \theta_i^{\alpha_i - 1}. 
\end{align}
After observing $x^*$, the posterior is given by 
\begin{align}
    p(\theta| x^* , \alpha) &\propto p(x^* | \theta)p(\theta | \alpha) \\
    &= \prod_i \theta_i^{\alpha_i - 1 + \mathbbm{1}_{x^{(i)}}(x^*)}. 
\end{align}
Note that this is again a Dirichlet distribution with hyperparameter $\alpha'$ such that 
\begin{align} \label{eq:true-update-dirichlet}
    \alpha'_i = \alpha_i + \mathbbm{1}_{x^{(i)}}(x^*), \ \ \  \forall i \in \{1,\ldots,n\}. 
\end{align}
That is, the hyperparameter corresponding to the observation $x^*$ is increased by one. This will make this observation more likely in the updated distribution.

\subsection*{Categorical model with latent variables} 

\subsubsection*{Exact posterior}

Now we let the model be a joint distribution over the product space of observations $x$ and latent states $z$, given by 
$\{x^{(1)}, \ldots,  x^{(n)}\} \times \{z^{(1)}, \ldots,  z^{(m)}\}$. 
The model is parametrized by $\theta = (\theta^D, \theta^A)$ as follows:
\begin{align} \label{eq:cat-dist1}
    p(z^{(j)}| \theta^D) = \theta^D_j, \\
    p(x^{(i)} | z^{(j)} , \theta^A) = \theta^A_{ij}. \label{eq:cat-dist2}
\end{align}
The prior over the $\theta$ is defined as follows:
\begin{align}
    p(\theta | \alpha) &= p(\theta^D | \alpha^D) \prod_j p(\theta^A_{\bullet j} | \alpha^A) \label{eq:prior-dir1}\\
    p(\theta^D | \alpha^D) &\propto  \prod_j \left(\theta^D_j\right)^{\alpha^D_j - 1},  \label{eq:prior-dir2}\\
    p(\theta^A_{\bullet j} | \alpha^A) &\propto  \prod_i \left(\theta^A_{ij}\right)^{\alpha^A_{i j} - 1},  \label{eq:prior-dir3}
\end{align}
where $\theta_{\bullet j}$ denotes the vector $(\theta_{1j},\ldots, \theta_{nj})$. After observing $x^* = x^{(i^*)}$, the exact posterior is given by 
\begin{align}
    p(\theta| x^* , \alpha) &\propto p(x^* | \theta)p(\theta | \alpha) \\
    &= \sum_j p(x^* | z^{(j)} , \theta^A) p(z^{(j)}| \theta^D) p(\theta^A | \alpha^A) p(\theta^D | \alpha^D) \\
    &\propto \sum_j \theta^A_{i^* j} \theta^D_j \left(\prod_{j'}  \prod_{i'} \left(\theta^A_{i'j'}\right)^{\alpha^A_{i' j'} - 1}\right) \left( \prod_{j''} \left(\theta^D_{j''}\right)^{\alpha^D_{j''} - 1}\right)\\
    &= \sum_j  \left(\prod_{j'}  \prod_{i'} \left(\theta^A_{i'j'}\right)^{\alpha^A_{i' j'} - 1 + \mathbbm{1}_{i^*}(i') \mathbbm{1}_{j}(j') } \right) \left(\prod_{j''} \left(\theta^D_{j''}\right)^{\alpha^D_{j''} - 1 +  \mathbbm{1}_{j}(j'')}\right). \label{eq:posterior-dirichlet-joint}
\end{align}
Note that this is no longer a Dirichlet distribution. Below we discuss how the Dirichlet distribution shows up in an algorithm for approximating the true posterior.  

\subsubsection*{Mean-field approximation}
We can also approximate the posterior over $\theta$ by minimizing the following variational free energy function:
\begin{align} \label{eq:vfe-cavi}
    F(\tilde{q}| x^*) = \bE_{\tilde{q}(z, \theta)}\left[ \ln \tilde{q}(z, \theta)   - \ln p(x^*, z, \theta | \alpha)\right],
\end{align}
where we assume the approximate posterior over both $\theta$ and $z$ factorizes as follows:
\begin{align}
    q(z, \theta) = q(z) q(\theta).
\end{align}

The coordinate ascent variational inference (CAVI) algorithm~\cite{blei2017variational, bishop2006pattern} updates the distributions $q(z)$ and $q(\theta)$ iteratively, each time holding one distribution fixed while updating the other.
More specifically, we can initialize $q(\theta)$ with our prior belief $p(\theta|\alpha)$ and minimize $F$ w.r.t.\ $\tilde{q}(z)$. The variational free energy now becomes
\begin{align} \label{eq:approx-post-learning1}
    F(\tilde{q}| x^*, q(\theta)) &= \bE_{\tilde{q}(z)}\left[ \ln \tilde{q}(z)  - \bE_{{q}(\theta)}[ \ln p(z| x^*, \theta)] \right] + C_{\setminus z}, 
\end{align}
where $C_{\setminus z}$ is independent of $\tilde{q}(z)$. The minimizer $q(z)$ is proportional to 
\begin{align} \label{eq:approx-post-learning2}
    q(z) \propto \exp \left( \bE_{{q}(\theta)}[ \ln p(z | x^*, \theta)] \right).
\end{align}
(See equation \eqref{eq:exactSolution}.) We then fix this $q(z)$ and optimize $F$ w.r.t.\ $\tilde{q}(\theta)$ and get 
\begin{align} \label{eq:approx-post-learning3}
    F(\tilde{q}| x^*, q(z)) &= \bE_{\tilde{q}(\theta)}\left[ \ln \tilde{q}(\theta)  - \bE_{{q}(z)}[ \ln p( x^*, z, \theta| \alpha)] \right] + C_{\setminus \theta},
\end{align}
where $C_{\setminus \theta}$ is independent of $\tilde{q}(\theta)$. The minimizer $q(\theta)$ is proportional to 
\begin{align} \label{eq:approx-post-learning4}
    q(\theta) &\propto \exp \left( \bE_{{q}(z)}[ \ln p( x^*, z, \theta | \alpha)] \right).
\end{align}

These two steps are performed iteratively until the beliefs about $\theta$ and $z$ have converged. 

Note that when $p(x, z, \theta | \alpha)$ is a categorical model with Dirichlet priors, as defined in \eqref{eq:cat-dist1}--\eqref{eq:prior-dir3}, 
\noeqref{eq:cat-dist2} \noeqref{eq:prior-dir1} \noeqref{eq:prior-dir2} 
we can rewrite equation \eqref{eq:approx-post-learning4} as follows: 
\begin{align}
    q(\theta) &\propto \exp \left( \bE_{{q}(z)}[\ln p( x^*, z, \theta|\alpha)] \right) \\
    &=  \exp \left( \bE_{{q}(z)} \left[ \ln p( x^*| z, \theta^A) + \ln p(z| \theta^D) + \ln p(\theta|\alpha) \right] \right) \\
    &= \exp \left( \sum_j q(z^{(j)}) \left[ \ln p( x^*| z^{(j)}, \theta^A) + \ln p(z^{(j)}| \theta^D) \right] \right) p(\theta|\alpha) \\
    &=  \prod_j \left(\theta^A_{i^*j} \right)^{q(z^{(j)})} \left(\theta^D_{j} \right)^{q(z^{(j)})}  \left(\theta^D_j\right)^{\alpha^D_j - 1} \prod_i \left(\theta^A_{ij}\right)^{\alpha^A_{i j} - 1}\\
    &=  \prod_j \left(\theta^D_j\right)^{\alpha^D_j + q(z^{(j)}) - 1} \prod_i \left(\theta^A_{ij}\right)^{\alpha^A_{i j} + \mathbbm{1}_{i^*}(i) q(z^{(j)}) - 1}
\end{align}
Note that this is again a Dirichlet distribution with updated parameter $\alpha'=({\alpha^D}', {\alpha^A}')$ given by
\begin{align} \label{eq:update-dirichlet-latent1}
    {\alpha_j^D}' &= \alpha^D_j + q(z^{(j)}),\\
    {\alpha_{ij}^A}' &= \alpha^A_{ij} + \mathbbm{1}_{i^*}(i) q(z^{(j)}). \label{eq:update-dirichlet-latent2}
\end{align}
Update \eqref{eq:update-dirichlet-latent1} makes latent states  with high $q(z)$ more likely in the updated distribution. Update \eqref{eq:update-dirichlet-latent2} makes sure that latent states are more likely to generate the observation $x^{(i^*)}$, especially those with high $q(z)$. 
Note the similarity with the update rule \eqref{eq:true-update-dirichlet} for categorical models without latent variables.  

\subsection*{POMDP model}

Now we let $p$ be the generative model from an active inference agent as described in \eqref{gen-mod1}--\eqref{gen-mod-prior4}.
\noeqref{gen-mod2} \noeqref{gen-mod3} \noeqref{gen-mod-prior1} \noeqref{gen-mod-prior2} \noeqref{gen-mod-prior3}
Similar to \eqref{eq:vfe-cavi} the variational free energy now becomes
\begin{multline}   \label{eq:vfe-learning}
    F(\tilde{q}| o_{1:T}, a_{1:T-1}) = \\
    \bE_{\tilde{q}(s_{1:T}, \theta)} \left[  \ln \tilde{q}(s_{1:T}, \theta) - \ln p(o_{1:T}, s_{1:T}, \theta| a_{1:T-1}, \alpha) \right].
\end{multline} 
We use the mean-field approximation $q(s_{1:T}, \theta) = q(s_{1:T}) q(\theta)$. Equivalent to \eqref{eq:approx-post-learning1}--\eqref{eq:approx-post-learning2} we can start by minimizing $F$ w.r.t.\ $\tilde{q}(s_{1:T})$ and get a minimizer $q_T(s_{1:T})$ using the current belief about $\theta$. Then, similar to \eqref{eq:approx-post-learning3}--\eqref{eq:approx-post-learning4}  we can use this minimizer to write $F$ as follows:
\begin{multline}  \label{eq:vfe-learning-fixed-qs2}
    F(\tilde{q}| o_{1:T}, a_{1:T-1}, q_T(s_{1:T})) = \\ \bE_{\tilde{q}(\theta)} \big[  
        \ln \tilde{q}(\theta) - \bE_{q_T(s_{1:T})} [\ln p(o_{1:T}, s_{1:T}, \theta| a_{1:T-1}, \alpha)]  \big] + C_{\setminus \theta},
\end{multline}
and work out the minimizer. This gives
\begin{align}
    q(\theta) &\begin{multlined}
        \propto \exp \left( \bE_{q_T(s_{1:T})} [\ln p(o_{1:T}, s_{1:T}, \theta| a_{1:T-1}, \alpha)] \right)
    \end{multlined} \\
    &\begin{multlined}
        = \exp \left( \bE_{q_T(s_{1:T})} \left[\sum_{\tau=1}^T \ln p(o_\tau |s_\tau, \theta^A ) + \ln p(s_1 | \theta^D)  \right. \right. \\
        \left. \left.  +\sum_{\tau=2}^T \ln p(s_{\tau} | s_{\tau-1}, a_{\tau-1}, \theta^B) \right] \right) p(\theta| \alpha)
    \end{multlined}\\
    &\begin{multlined}
        = \exp \left( \sum_{s_{1:T}} q_T(s_{1:T}) \left[\sum_{\tau=1}^T \ln p(o_\tau |s_\tau, \theta^A ) + \ln p(s_1 | \theta^D)  \right. \right. \\
        \left. \left.  +\sum_{\tau=2}^T \ln p(s_{\tau} | s_{\tau-1}, a_{\tau-1}, \theta^B) \right] \right) p(\theta| \alpha)
    \end{multlined}\\
    &\begin{multlined}
        =   \prod_j \left(\theta^D_{j} \right)^{q_T(s_1^{(j)})} \prod_i \left(\theta^A_{ij} \right)^{\sum_{\tau=1}^T \mathbbm{1}_{o^{(i)}}(o_\tau) q_T(s_\tau^{(j)})}    \\
        \prod_k \left(\theta_{jkl}^B\right)^{\sum_{\tau=2}^{T} q_T(s_{\tau}^{(j)}) q_T(s_{\tau-1}^{(k)} )  \mathbbm{1}_{a^{(l)}}(a_{\tau-1})} p(\theta| \alpha)
    \end{multlined}\\
    &\begin{multlined}
        =   \prod_j \left(\theta^D_{j} \right)^{\alpha^D_j + q_T(s_1^{(j)}) - 1} \prod_i \left(\theta^A_{ij} \right)^{\alpha^A_{ij} + \sum_{\tau=1}^T \mathbbm{1}_{o^{(i)}}(o_\tau) q_T(s_\tau^{(j)}) - 1}    \\
        \prod_k \left(\theta_{jkl}^B\right)^{\alpha_{jkl}^B + \sum_{\tau=2}^{T} q_T(s_{\tau}^{(j)}) q_T(s_{\tau-1}^{(k)} )  \mathbbm{1}_{a^{(l)}}(a_{\tau-1}) - 1}. 
    \end{multlined}\\
\end{align}
This gives the update rules for $\alpha$ given in \eqref{eq:approx-update-D}--\eqref{eq:approx-update-B}. 

\begin{remark}
    Note that these update rules are actually just the first iteration of the CAVI algorithm described above. It will therefore in general not minimize the variational free energy in equation \eqref{eq:vfe-learning}. 
    Instead it minimizes the quantity given in equation \eqref{eq:vfe-learning-fixed-qs1} and \eqref{eq:vfe-learning-fixed-qs2} where $q_T(s_{1:T})$ is fixed. Note however that if one would assume $q_T(s_{1:T})$ to be given, the following variational free energy would be the natural choice to minimize: 
    \begin{align}  
        \bE_{\tilde{q}(\theta)} \big[  
            \ln \tilde{q}(\theta) - \ln \left(\bE_{q_T(s_{1:T})} [p(o_{1:T}, \theta| s_{1:T}, a_{1:T-1}, \alpha)]\right)  \big],
    \end{align}
    since this has as minimizer the exact posterior distribution. 
\end{remark}

\section{Further details}
\label{further-details}

\subsubsection*{Equivalent formulations of expected free energy} 
Recall that the expected free energy in equation \eqref{eq:efe} is given by
\begin{align}  \label{eq:efe1App}
    \begin{aligned}
        G(\pi_t| o_{1:t}, a_{1:t-1}) =  - \bigg(&\bE_{q_{t}(o_{t+1:T}| \pi_t)}\left[\DKL \Big( q_{t}(s_{t+1:T} | o_{t+1:T}, \pi_t) \parallel q_{t}(s_{t+1:T}|\pi_t) \Big) \right] \\ 
    &+ \bE_{q_{t}(o_{t+1:T} | \pi_t)} \Big[ \ln p_C(o_{t+1:T})\Big]\bigg).
    \end{aligned}
\end{align}
We can derive the equivalent formulation from equation \eqref{eq:efe2} as follows. We first expand the KL divergence term to get
\begin{align}  \label{eq:efeI1}
    \begin{aligned}
        G(\pi_t| o_{1:t}, a_{1:t-1}) =  \bE_{q_{t}(s_{t+1:T}, o_{t+1:T}| \pi_t)}\Big[ &\ln  q_{t}(s_{t+1:T}|\pi_t) - \ln q_{t}(s_{t+1:T} | o_{t+1:T}, \pi_t)  \\ 
        &- \ln  p_C(o_{t+1:T})\Big].
    \end{aligned}
\end{align}
Using Bayes' rule we can rewrite 
\begin{align}
\begin{aligned}
    - \ln q_{t}(s_{t+1:T} |o_{t+1:T}, \pi_t) = &- \ln q_{t}(o_{t+1:T} |s_{t+1:T}, \pi_t) - \ln q_{t}(s_{t+1:T} |\pi_t) \\
    &+ \ln q_{t}(o_{t+1:T} |\pi_t).
\end{aligned}
\end{align}
Plugging this into \eqref{eq:efeI1} and using that $q_{t}(o_{t+1:T} |s_{t+1:T}, \pi_t) = p(o_{t+1:T} |s_{t+1:T})$ we get 
\begin{align}
    &\begin{aligned}
        G(\pi_t| o_{1:t}, a_{1:t-1}) =  \bE_{q_{t}(s_{t+1:T}, o_{t+1:T}| \pi_t)}\Big[ &\ln  p(o_{t+1:T}|s_{t+1:T}) + \ln q_{t}(o_{t+1:T} | \pi_t)  \\ 
        &- \ln p_C(o_{t+1:T})\Big]
    \end{aligned}\\
    &\begin{aligned}
        \phantom{G(\pi_t| o_{1:t}, a_{1:t-1})} = \ &\bE_{q_{t}(s_{t+1:T} | \pi_t)}\Big[\mathrm{H}\big[p(o_{t+1:T} |s_{t+1:T})\big]\Big] \\ 
        & + \DKL\Big(q_{t}(o_{t+1:T} | \pi_t) \parallel p_C(o_{t+1:T})\Big),
    \end{aligned} \label{eq:efe2App}
\end{align}
where the last line is equal to equation \eqref{eq:efe2}.

\subsubsection*{Independence between state factors and observation modalities}

In order to make the computation of state inference more efficient, the agent can use independencies between different state factors and observation modalities (different dimensions of state and observation space). More specifically, we assume that given a state, the different observation modalities are independent, which translates to:
\begin{align}
    p(o_\tau | s_\tau) = \prod_m p(o_\tau^m | s_\tau).
\end{align}
We use superscript $m$ and $f$ to denote a specific observation modalities and state factors respectively. Furthermore, we assume a certain state factor to be independent of all other state factors in the same and previous time step, given the same state factor in the previous time step and the last action, i.e.:
\begin{align}
    p(s_\tau|s_{\tau-1}, a_{\tau-1}) = \prod_f p(s_\tau^f|s_{\tau-1}^f, a_{\tau-1}).
\end{align}
For a fixed state factor $f$ equation \eqref{eq:update-state} and \eqref{eq:posterior-future-state} now become
\begin{align} \label{eq:update-state-factor}
    q_{t}(s_t^f) &\propto p(o_t|s_t^f) \sum_{s^f_{t-1}} p(s_t^f|s_{t-1}^f, a_{t-1}) q_{t-1}(s_{t-1}^f)\\
    q_{t}(s^f_\tau| a_{1:\tau-1} ) &= \sum_{s^f_{\tau-1}} p(s^f_\tau| s^f_{\tau-1}, a_{\tau-1})  q_{t}(s^f_{\tau-1}| a_{1:\tau-2}), 
\end{align}
and equation \eqref{eq:posterior-states-past-future} becomes
\begin{align}
    q_{t}(s_{1:T}|\pi_t) &\propto \prod_{\tau=1}^t \prod_m p(o_\tau^m | s_\tau) \ p(s_1) \prod_{\tau=2}^T \prod_f p(s_\tau^f|s_{\tau-1}^f, a_{\tau-1}). 
\end{align}

\subsubsection*{Fixed point iteration}
Equation \eqref{eq:update-state-factor} involves the distribution $p(o_t|s_t^f)$. We can however not access this directly. To find an approximate solution, we can use fixed point iteration as follows:
\begin{align}
    q_{t}^{(i+1)}(s_t^f) &\propto \sum_{s^{\setminus f}_t} q_{t}^{(i)}(s_t^{\setminus f}) p(o_t|s_t) \sum_{s^f_{t-1}} p(s_t^f|s_{t-1}^f, a_{t-1}) q_{t-1}(s_{t-1}^f),
\end{align}
where $\setminus f$ denotes the set of all state factors apart from $f$.

\subsubsection*{Softmax function}

\begin{definition}
    Let $\cS=\{x^{(1)},\ldots,x^{(n)}\}$ be a finite set and $\mu: \cS \to \bR$ a function. The \emph{softmax function} $\sigma$ is given by 
    \begin{align} \label{eq:softmax-function}
        \sigma(\mu(x^{(i)})) = \frac{e^{\mu(x^{(i)})}}{\sum_j e^{\mu(x^{(j)})}}.
    \end{align}
\end{definition}
Note that $\sigma(\mu(x^{(i)})) > 1$ and $\sum_\cS \sigma(\mu(x^{(i)}))=1$. Therefore the softmax function can be used to turn $\mu$ into a probability distribution.

\end{document}